# An Efficient Triplet-based Algorithm for Evidential Reasoning

Yaxin Bi
School of Computing and Mathematics
University of Ulster at Jordanstown
Co. Antrim, BT37 0QB, UK

Jiwen Guan
School of Computer Science
Queen's University of Belfast
Belfast, BT7 1NN, UK

## Abstract

Linear-time computational techniques have been developed for combining evidence which is available on a number of contending hypotheses. They offer a means of making the computation-intensive calculations involved more efficient in certain circumstances. Unfortunately, they restrict the orthogonal sum of evidential functions to the *dichotomous structure* − applies only to elements and their complements. In this paper, we present a novel evidence structure in terms of a *triplet* and a set of algorithms for evidential reasoning. The merit of this structure is that it divides a set of evidence into three subsets, *distinguishing trivial evidential elements from important ones* − focusing some particular elements. It avoids the deficits of the dichotomous structure in representing the preference of evidence and estimating the basic probability assignment of evidence. We have established a formalism for this structure and the general formulae for combining pieces of evidence in the form of the triplet, which have been theoretically and empirically justified.

## 1 INTRODUCTION

In this paper, we propose a novel evidence structure − a *triplet* − which can be used to represent multiple pieces of evidence to improve effectiveness and efficiency in computing Dempster's rule of combination and in making decisions. In particular, it is designed to represent the outputs of ensemble classifiers and it is capable of incorporating the prioritized outputs − evidence − into the decision making process, resulting in an improvement of classification accuracy. We have developed a formalism for the triplet structure, and we show that the mass functions defined on this formalism satisfy the properties embedded in the Dempster-Shafer theory of evidence (Shafer 1976). We also developed a range of formulae for realizing linear time computations of combining various triplet mass functions, which are applicable to the other evidential functions of *belief function*, *commonality function*, *plausibility function*, and *doubt function*.

The evidential functions are defined on sets and their enumeration. Broadly speaking, the time complexity of computing evidential functions is exponential. Essentially, it is not feasible to translate the evidence theory into an efficient direct implementation (Barnett 1981). Therefore the applications of the Dempster-Shafer theory of evidence to real-world problems have been until recently limited (Denoeux 2000), in particular, to text categorization. Considerable efforts have been devoted to developing efficient algorithms for computing evidential functions (Barnett 1981; Shafer, et al. 1987; Kennes, et al. 1990; Moral, et al. 1994; Guan, et al. 1995; Shafer, et al. 2002). However, a particular structure embedded in evidence plays a dominant role in developing practical and linear algorithms for computing evidential functions and for real-world applications (Xu, et al. 1992; Denoeux 2000). For example, Barnett (1981) proposed a method for partitioning an evidence space in several independent ways and clustering pieces of evidence into the dichotomous partitions of a singleton proposition and its negation. Based on the dichotomous structure, a range of formulae were developed for computing various evidential functions, in which the computations of various evidential functions can be performed in linear time (Barnett 1981; Shafer, et al. 1987; Guan, et al. 1995).

Barnett's linear-time technique provides a feasible way for efficiently computing various evidential functions. However, two issues arise when this technique is applied to combining the outputs of ensemble classifiers. The first is that given a list of evidence, the dichotomous structure does not distinguish trivial evidential elements from important ones. This could lead to a deterioration in the performance of classification decision when multiple classifiers are combined. The second is that it is difficult to establish a measure for obtaining basic probability assignments for dichotomous elements − *a single focus, its complement*, and *ignorance*. In practice, when multiple pieces of evidence of classifiers being combined in a multiple classifier system, evidence elements (decision classes) consist of a list of decision classes. These decision classes are not equally important in determining the final decision. It is necessary to introduce a weighting mechanism to give high weights to decisions deemed important and lower weights to less important ones, and incorporate these weights into the decision making process to determine the final decision.

The awareness of the above issues motivates us to develop a novel *triplet* structure to cope with the above issues. To obtain a triplet from a list of decision classes, we have developed a new operation, called an *outstanding rule* − a focusing operation. It not only separates trivial elements from important ones, but also serves as an effective mechanism for allocating basic probability assignments to focal elements within triplets. To justify the triplet structure and the outstanding rule, we provide a theoretical support for how this rule can be used to derive mass functions and establish the formulae for combining two or more results obtained by the outstanding rule. The computational complexity of these formulae can be achieved in linear time (Bi 2004).

## 2 DEMPSTER-SHAFER (DS) THEORY OF EVIDENCE

Given a frame of discernment consisting of mutually exclusive propositions, for any subset $H \subseteq \Theta$ or $H \in 2^\Theta$, called a focal element or focus. It is also called a singleton when $H$ is a one element subset. The DS theory uses a numeric value in an interval $[0, 1]$ to represent the degree of support for the subset $H$, denoted by $m(H)$, called a *mass function* (Shafer 1975).

**Definition 1** Let $m_1$ and $m_2$ be two mass functions on the frame of discernment $\Theta$, and for any subset $H \subseteq \Theta$, the *orthogonal sum* of two mass functions on $H$ is defined as:

$$m_1 \oplus m_2(H) = \frac{\sum_{X \cap Y = H} m_1(X) m_2(Y)}{1 - \sum_{X \cap Y = \emptyset} m_1(X) m_2(Y)} \quad (1)$$

The orthogonal sum is also called Dempster's rule of combination. It is an important operation, allowing two mass functions to be combined into a third mass function.

### 2.1 THE DICHOTOMOUS FUNCTION

In particular, a mass function $m$ is said to be a dichotomous function if the only possible focal elements of $m$ are $A, \Theta - A, \Theta$ for some $A \subseteq \Theta$. A special case occurs when $A$ is a singleton. In such a situation, a dichotomous mass function $m$ has no focuses other than $\{x\}, \Theta - \{x\}, \Theta$ for some $x$ which is referred to as a dichotomous structure (Barnett 1981).

Let $\Theta = \{x_1, x_2, ..., x_{|\Theta|}\}$. Suppose that for every $i = 1, 2, ..., |\Theta|$, there is a dichotomous mass function $m_i$: $p_i = m_i(\{x_i\}), c_i = m_i(\Theta - \{x_i\}), r_i = m_i(\Theta)$, where $p_i + c_i + r_i = 1$ for $i = 1, 2, ..., |\Theta|$. We view these quantities as follows:

- $p_i$ is the measure of support for $\{x_i\}$;
- $c_i$ is the measure of support for the refutation of $\{x_i\}$; and
- $r_i$ is the measure of the support not assigned for or against the proposition $\{x_i\}$.

Barnett's technique is based on the dichotomous mass functions instead of general mass functions. It means that instead of potentially exponential calculation, the computation of dichotomous mass functions involves only the 3 particular subsets $\{x\}, \Theta - \{x\}, \Theta$ for each $x \in \Theta$, while the general mass functions have to enumerate all $2^{|\Theta|}$ subsets of $\Theta$.

Barnett's approach is to consider the *entire* orthogonal sum for evidence bodies which have the structure $m_1 \oplus m_2 \oplus ... \oplus m_{|\Theta|}$. These are precisely those evidence spaces which are separable into exactly $|\Theta|$ dichotomous mass functions $m_1, m_2, ..., m_{|\Theta|}$. Guan, et al. (1995) generalized this to consider the general orthogonal sum as explained below.

Let $\Theta = \{x_1, ..., x_{|\Theta|}\}$. Suppose that for some $x_i \in \Theta$, there are $l_i$ dichotomous mass functions of repeat focuses as defined previously: $m_{ij}(\{x_i\}) = p_{ij}, m_{ij}(\Theta - \{x_i\}) = c_{ij}, m_{ij}(\Theta) = r_{ij}; p_{ij} + c_{ij} + r_{ij} = 1$; where $i = 1, 2, ..., |\Theta|$; $j = 1, 2, ..., l_i$. The task now is to calculate quantities associated with

$$m = \underbrace{m_{11} \oplus ... \oplus m_{1l_1}}_{l_1 \text{ terms}} ... \oplus \underbrace{m_{k1} \oplus ... \oplus m_{kl_k}}_{l_k \text{ terms}} \quad (2)$$

where $1 \leq k \leq |\Theta|$; $0 \leq l_1, l_2, ..., l_k$ and $l_1 + l_2 + + l_k = n$, and $n$ is the number of items to be summed which may be greater than $|\Theta|$. From Equation (2), the calculation of combining $n$ dichotomous mass functions is divided into two parts. The first part is to combine the mass functions with repeated focal elements in terms of repeated focuses. The second part is to combine the mass functions without repeated focuses. Here we only present the former to show that the calculation of combining such mass functions requires linear time (Guan, et al. 1995).

## 2.2 FORMULAE FOR REPEATED FOCUSES

To establish the computational formulae for the case where there are repeated focuses, there is a need to prove the combinability of dichotomous mass functions. In other words, we need to show the normalization constant $K^{-1} \neq 0$.

Suppose that we have $l$ dichotomous mass functions with repeated focuses: $m_i(\{x_i\}) = p_i, m_i(\Theta - \{x_i\}) = c_i, m_i(\Theta) = r_i; p_i + c_i + r_i = 1$; where $i = 1, 2, ..., l$. and $l$ is an arbitrary integer. The task now is to calculate quantities associated with $m_1 \oplus m_2 \oplus ... \oplus m_l$. By definition the orthogonal sum, a collection of mass functions is given by

$$m(A) = K \sum_{A_1 \cap ... \cap A_l = A} m_1(A_1)...m_l(A_l) \quad (3)$$

where $K^{-1} = \sum_{A_1 \cap ... \cap A_l \neq \emptyset} m_1(A_1)m_2(A_2)...m_l(A_l)$. To prove $K^{-1} \neq 0$, the formulae for computing $K$ were established as follows (Guan, et al. 1995). Let $m_1, m_2, ...m_l$ be $l$ dichotomous mass functions with the same focuses $\{x\}, \Theta - \{x\}, \Theta$, which have the following conditions:

$$p_i = m_i(\{x\}), c_i = m_i(\Theta - \{x\}), r_i = m_i(\Theta) \quad (4)$$

$$0 \leq p_i, c_i, r_i \leq 1; p_i + c_i + r_i = 1 \quad (5)$$

$$d_i = 1 - p_i = c_i + r_i \quad (6)$$

Then we have

$$K^{-1} = \sum_{A_i \subseteq \Theta, \cap_{i=1,...,l} A_i \neq \emptyset} m_1(A_1)...m_l(A_l)$$

$$= p_1 \prod_{i=2}^{l}(p_i + r_i) + c_1 \prod_{i=2}^{l}(c_i + r_i)$$

$$+ r_1 p_2 \prod_{i=3}^{l}(p_i + r_i) + r_1 c_2 \prod_{i=3}^{l}(c_i + r_i) \quad (7)$$

$$+ r_1 r_2 p_3 \prod_{i=4}^{l}(p_i + r_i) + r_1 r_2 c_3 \prod_{i=4}^{l}(c_i + r_i) + ... + \prod_{i=1}^{l} r_i.$$

There are two special cases: 1) if $p_i = 1$ for some $i$, then $K^{-1} = \prod_{j \neq i; j=1,...,l}(p_j + r_j) = \prod_{j \neq i; j=1,...,l}(1 - c_j)$; 2) if $c_i = 1$ for some $i$, then $K^{-1} = \prod_{j \neq i; j=1,2,...,l}(c_j + r_j). = \prod_{j \neq i; j=1,2,...,l} d_j$.

Thus the dichotomous mass functions $m_1, m_2, ..., m_l$ are combinable if and only if $K^{-1} \neq 0$, whose proof has been given (Guan, et al. 1995). From Equation (7), it is shown that the computation of combining dichotomous mass functions is achieved in linear time. In fact, a straightforward interpretation on the computational efforts of formula (7) is that a number of computations increase linearly with the number of focal elements in the frame of discernment (Shafer, et al. 1987).

## 2.3 COMBINING DICHOTOMOUS FUNCTIONS

According to the combinability of dichotomous mass functions, the formulae of combining evidential functions with repeated focuses are established below. Let $m_1, m_2, ..., m_l$ be $l$ dichotomous mass functions with the same focuses on $2^\Theta$: $p_i = m_i(\{x\}), c_i = m_i(\Theta - \{x\}), r_i = m_i(\Theta), m_i(elsewhere) = 0, i = 1, 2, ..., l$. If $K^{-1} > 0$, the orthogonal sum $m = m_1 \oplus, ..., \oplus m_l; 1 \leq l \leq |\Theta|$ of $l$ dichotomous mass functions $m_1, m_2, ..., m_l$ is again dichotomous with the same focuses. The formula for computing $m$ on $2^\Theta$ is as follows: $m(\emptyset) = 0$,

$$m(\{x\}) = p = K(p_1 \prod_{i=2}^{l}(p_i + r_i) + r_1 p_2 \prod_{i=3}^{l}(p_i + r_i)$$

$$+ r_1 r_2 p_3 \prod_{i=4}^{l}(p_i + r_i) + ... + r_1 r_2 ... r_{l-2} p_{l-1}(p_l + r_l)), \quad (8)$$

$$m(\Theta - \{x\}) = c = K(c_1 \prod_{i=2}^{l}(c_i + r_i) + r_1 c_2 \prod_{i=3}^{l}(c_i + r_i)$$

$$+ r_1 r_2 c_3 \prod_{i=4}^{l}(c_i + r_i) + ... + r_1 r_2 ... r_{l-2} c_{l-1}(c_l + r_l)), \quad (9)$$

$$m(\Theta) = r = K \prod_{i=1}^{l} r_i, m(elsewhere) = 0; \quad (10)$$

Here, we only present a review of the combinability and the calculation formulae of the dichotomous functions based on the case where the focal elements are repeated. More details for the case of different focuses have been presented by Guan, et al. (1995).

## 3 TRIPLET AND OUTSTANDING RULE

In this section, we start with a formalism of a new evidence structure, and then we prove that our techniques are compatible with the conventional evidential function − mass function.

**Definition 2** Let $\Theta$ be a frame of discernment and $\varphi(d) = \{m(\{x_1\}), m(\{x_2\}), ..., m(\{x_n\})\}$, where $|n| \geq$

2, an expression of the form $Y = \langle A_1, A_2, A_3 \rangle$ is defined as a *triplet*, where $A_1, A_2$ are singletons, and $A_3$ is the whole set $\Theta$, and they satisfy

$$m(A_1) \oplus m(A_2) \oplus m(A_3) = 1$$

These elements are obtained by using a focusing operation $\sigma$ on $m$, called the *outstanding rule* and denoted by $m^\sigma$ as follows:

$$A_1 = arg \max m(\{x_1\}), m(\{x_2\}), ..., m(\{x_n\}) \quad (11)$$

$$A_2 = arg \max m(\{x\} \mid x \in \{x_1, ..., x_n\} - \{u\}) \quad (12)$$

$$A_3 = \Theta, m^\sigma(\Theta) = 1 - m^\sigma(\{u\}) + m^\sigma(\{v\}) \quad (13)$$

We refer to $m^\sigma$ as a *triplet mass function* or as a *two-point mass function*. Likewise, we can consider mass functions for *three-point focuses, four-point focuses, ..., n-point focuses*. For the sake of simplicity, here we only consider mass functions for two-point focuses and their use. In comparison with the dichotomous structure, the triplet structure has the following properties:

- When a frame of discernment $\Theta$ contains two focal elements, a triplet function is the same as a dichotomous function.

- A dichotomous structure only has one variable, so if any singleton proposition or its negation is known, then the dichotomous structure is determined. However a triplet structure involves two variables, so the triplet structure is more general and complicated than the dichotomous structure.

- To carry out the combination of two triplet functions, we look at the configuration of any two pairs of focal subsets, $A_1, A_2$ and $B_1, B_2$. For *dichotomous functions*: 1) If $A_1 = B_1$ then $A_2 = B_2$, $A_1 \cap B_2 = \emptyset$ and $A_2 \cap B1 = \emptyset$, so the combined result involves three different focal elements. 2) If $A_1 \neq B_1$ then $A_2 \neq B_2$, $A_1 \cap B_2 \neq \emptyset$ and $A_2 \cap B_1 \neq \emptyset$, so the combined result involves five different focal elements. For *triplet functions*: 1) If $A_1 = B_1$ and $A_2 = B_2$, then $A_1 \cap B_2 = \emptyset$ and $A_2 \cap B_1 = \emptyset$, so the combination of two triplet functions involves three different focal elements. 2) If $A_1 = B_1$ and $A_2 \neq B_2$ then $A_1 \cap B_2 = \emptyset, A_2 \cap B_1 = \emptyset$ and $A_2 \cap B_2 = \emptyset$ or if $A_2 = B_2$ and $A_1 \neq B_1$, then $A_1 \cap B_2 = \emptyset, A_2 \cap B_1 = \emptyset$ and $A_1 \cap B_1 = \emptyset$, so the combination of two triplet functions involves four different focal elements. 3) If $A_1 \neq B_1, A_2 \neq B_2, A_1 \neq B_2$, and $A_2 \neq B_1$ then $A_1 \cap B_1 = \emptyset, A_2 \cap B_2 = \emptyset, A_1 \cap B_2 = \emptyset$ and $A_2 \cap B_1 = \emptyset$, so the combination involves five different focal elements.

- Given triplet functions and dichotomous functions, intuitively the computations of combining two triplet functions is more efficient than that of combining two dichotomous functions because the computation of dichotomous functions requires more set operations than triplet functions. An empirical comparison is illustrated in Section 5.

Through the above comparisons, it can be seen that the formulae for computing dichotomous functions are not directly applicable to triplet functions. In cases 2) and 3) of triplet functions, the combinations of multiple triplet functions cannot be iteratively performed since we are only interested in combining two-point focuses. In the following sections, we provide theoretical proofs to show any two triplet mass functions are indeed combinable and develop formulae for computing the combinations of triplet functions for three different cases above (Bi 2004).

### 3.1 TWO-POINT FOCUSES EQUAL

Considering the case where two focal singletons $\{x_1\}, \{y_1\}$ in one triplet are equal to $\{x_2\}, \{y_2\}$ in another triplet, i.e. $x_1 = x_2, y_1 = y_2 (x_1 \neq y_1)$, first, we need to show the combination of $m_1 \oplus m_2$ does exist and then establish formulae to compute their combination.

**Theorem 1** Let be a frame of discernment, $m_1, m_2$ be two triplet mass function on $\Theta$, and $\{x\}, \{y\}$ and $\{x\}, \{y\}(x \neq y)$ be two pairs of two-point focal elements with the condition of,

$$m_1(\{x\}) + m_1(\{y\}) + m_1(\Theta) = 1$$

$$m_2(\{x\}) + m_2(\{y\}) + m_2(\Theta) = 1$$

Then

$$K^{-1} = 1 - m_1(\{x\})m_2(\{y\}) - m_1(\{y\})m_2(\{x\}),$$

and $m_1, m_2$ are combinable if and only if

$$m_1(\{x\})m_2(\{y\}) + m_1(\{y\})m_2(\{x\}) < 1.$$

By the intersection table of $m_1$ and $m_2$, we can easily obtain a normalization factor $K^{-1}$ and then prove the above condition must be held when $m_1$ and $m_2$ are validly combined using the orthogonal sum operation.

This theorem reveals the combinability of triplet mass functions. Based on this, we establish the formulae for computing the combination of two triplet mass functions below:

$$(m_1 \oplus m_2)(\{x\}) = K(m_1(\{x\})m_2(\{x\}) \quad (14)$$

$$+ m_1(\{x\})m_2(\Theta) + m_1(\Theta)m_2(\{x\})),$$

$$(m_1 \oplus m_2)(\{y\}) = K(m_1(\{y\})m_2(\{y\}) \quad (15)$$
$$+ m_1(\{y\})m_2(\Theta) + m_1(\Theta)m_2(\{y\})),$$
$$(m_1 \oplus m_2)(\Theta) = Km_1(\Theta)m_2(\Theta), \quad (16)$$

where
$$K^{-1} = 1 - \sum_{X \cap Y = \emptyset} m_1(X)m_2(Y) \quad (17)$$

## 3.2 ONE TWO-POINT FOCUSES EQUAL

In this section, let us consider the case where given two triplet mass functions $m_1$ and $m_2$, a focal element in one triplet is equal to one in another triplet. Theorem 2 says that the two mass functions are combinable.

**Theorem 2** Let $\Theta$ be a frame of discernment, $m_1, m_2$ be two triplet mass function on $\Theta$, and also let $\{x\}, \{y\}$ and $\{x\}, \{z\}(y \neq z)$ be two pairs of two-point focal elements with the following condition:

$$m_1(\{x\}) + m_1(\{y\}) + m_1(\Theta) = 1$$
$$m_2(\{x\}) + m_2(\{z\}) + m_2(\Theta) = 1$$

Then
$$K^{-1} = 1 - m_1(\{x\})m_2(\{z\}) - m_1(\{y\})m_2(\{z\}) -$$
$$m_1(\{y\})m_2(\{x\}),$$

and $m_1, m_2$ are combinable if and only if the following constraint is held:

$$m_1(\{x\})m_2(\{z\}) + m_1(\{y\})m_2(\{z\})$$
$$+ m_1(\{y\})m_2(\{x\}) < 1.$$

Following Theorem 2, a new mass function can be obtained from the two individual triplet mass functions. Thus by the orthogonal sum rule, the general formulae for computing combinations of mass functions are given below:

$$(m_1 \oplus m_2)(\{x\}) = K(m_1(\{x\})m_2(\{x\})$$
$$+ m_1(\{x\})m_2(\Theta) + m_1(\Theta)m_2(\{x\})), \quad (18)$$
$$(m_1 \oplus m_2)(\{y\}) = Km_1(\{y\})m_2(\Theta), \quad (19)$$
$$(m_1 \oplus m_2)(\{z\}) = Km_1(\Theta)m_2(\{z\}), \quad (20)$$
$$(m_1 \oplus m_2)(\Theta) = Km_1(\Theta)m_2(\Theta), \quad (21)$$

where
$$K^{-1} = 1 - m_1(\{x\})m_2(\{z\})$$
$$- m_1(\{y\})m_2(\{z\}) - m_1(\{y\})m_2(\{x\}). \quad (22)$$

The new mass function $m_1 \oplus m_2$ is no longer a triplet mass function, and it now involves four different focal elements $\{x\}, \{y\}, \{z\}$, and $\Theta$. For more than two triplet functions, the combining process cannot proceed iteratively since we are only interested in two-point focuses. However, by applying the outstanding rule, the combined result can be transformed to a new triplet mass function. We detail the computational steps below.

By Definition 2, we have a new function $(m_1 \oplus m_2)^\sigma$ as follows:

$$(m_1 \oplus m_2)^\sigma(\{x'\}) + (m_1 \oplus m_2)^\sigma(\{y'\})$$
$$+ (m_1 \oplus m_2)^\sigma(\Theta) = 1.$$

To obtain $(m_1 \oplus m_2)^\sigma$, we assume
$$m_1 \oplus m_2(\{x\}) = f(x); m_1 \oplus m_2(\{y\}) = f(y);$$
$$m_1 \oplus m_2(\{z\}) = f(x).$$

Then for focal element $\{x'\}$ we have
$$m_1 \oplus m_2(\{x'\}) = f(x') \quad (23)$$

where $\{x'\} = arg \max(f(x), f(y), f(z))$.

For focal element $\{y'\}$ we have
$$m_1 \oplus m_2(\{y'\}) = f(y') \quad (24)$$

where $y' = arg \max(f(t)|t \in (\{x, y, z\} - \{x'\}))$.

For focal element $\Theta$ we have
$$(m_1 \oplus m_2)^\sigma(\Theta) = 1 - f(x') - f(y'). \quad (25)$$

## 3.3 COMPLETELY DIFFERENT TWO-POINT FOCUSES

Finally, let us examine the case where no focal element is in common in two triplets. As indicated previously, the combination of such two triplet mass functions will involve five different focal elements. We first prove such triplet functions are combinable.

**Theorem 3** Let $\Theta$ be a frame of discernment, let $m_1, m_2$ be two triplet functions, and $\{x\}, \{y\}$ and $\{u\}, \{v\}(x \neq y, x \neq u$ and $y \neq v)$ be two pairs of focal elements along with the following conditions:

$$m_1(\{x\}) + m_1(\{y\}) + m_1(\Theta) = 1,$$
$$m_2(\{u\}) + m_2(\{v\}) + m_2(\Theta) = 1.$$

Then
$$K^{-1} = 1 - \sum_{X \cap Y = \emptyset} m_1(X)m_2(Y) =$$
$$1 - m_1(\{x\})m_2(\{u\}) - m_1(\{x\})m_2(\{v\}) -$$
$$m_1(\{y\})m_2(\{u\}) - m_1(\{y\})m_2(\{v\}),$$

and $m_1, m_2$ are combinable if and only if the following constraint is held:

$$m_1(\{x\})m_2(\{u\}) + m_1(\{x\})m_2(\{v\})+$$

$$m_1(\{y\})m_2(\{u\}) + m_1(\{y\})m_2(\{v\}) < 1$$

It is not difficult to prove the above theorem on the basis of the orthogonal sum operation. By Theorem 3, we develop the formulae for computing each focal element below:

$$(m_1 \oplus m_2)(\{x\}) = Km_1(\{y\})m_2(\Theta) = f(x),$$

$$(m_1 \oplus m_2)(\{y\}) = Km_1(\{y\})m_2(\Theta) = f(y),$$

$$(m_1 \oplus m_2)(\{u\}) = Km_1(\Theta)m_2(\{z\}) = f(u),$$

$$(m_1 \oplus m_2)(\{v\}) = Km_1(\Theta)m_2(\{z\}) = f(v),$$

where

$$K^{-1} = 1 - \sum_{X \cap Y = \emptyset} m_1(X)m_2(Y) =$$

$$1 - m_1(\{x\})m_2(\{u\}) - m_1(\{x\})m_2(\{v\})-$$

$$m_1(\{y\})m_2(\{u\}) - m_1(\{y\})m_2(\{v\}),$$

However, the same situation occurs as in Section 3.2, the combination of $m_1, m_2$ is no longer a triplet mass function, it now involves five focal elements $\{x\}, \{y\}, \{u\}, \{v\}, \Theta$, therefore further combinations with more triplet functions are invalid in this context. Likewise, to obtain a new triplet mass function, there is a need to apply the outstanding rule to the combined result.

More specifically, by Definition 2, we can obtain a new function $(m_1 \oplus m_2)^\sigma$ as follows:

$$(m_1 \oplus m_2)^\sigma(\{x'\}) + (m_1 \oplus m_2)^\sigma(\{y'\})$$

$$+(m_1 \oplus m_2)^\sigma(\Theta) = 1.$$

Then for focal element $\{x'\}$ we have

$$m_1 \oplus m_2(\{x'\}) = f(x') \qquad (26)$$

where $\{x'\} = arg\max(f(x), f(y), f(u), f(v))$.

For focal element $\{y'\}$ we have

$$m_1 \oplus m_2(\{y'\}) = f(y') \qquad (27)$$

where $y' = arg\max(f(t)|t \in (\{x,y,u,v\} - \{x'\}))$.

Finally, for focal element $\Theta$ we have

$$(m_1 \oplus m_2)^\sigma(\Theta) = 1 - f(x') - f(y'). \qquad (28)$$

## 4 AN APPROXIMATE FORMULA FOR COMBINING MORE THAN TWO TRIPLET FUNCTIONS

We have shown that any two triplet functions are combinable and we have also established the formulae for computing the combinations of two triplet functions. By repeatedly applying these formulae, we combine any number of triplet functions. In order to combine a collection of triplet mass functions $m_1, m_2, ..., m_n$, we can simply form pairwise orthogonal sums as follows:

$$m_1 \oplus m_2$$

$$(m_1 \oplus m_2) \oplus m_3$$

$$((m_1 \oplus m_2) \oplus m_3) \oplus m_4 \oplus ...$$

this continues until all $m_i$ are included. As proved above, at each stage of this process, each combined result is a triplet mass function that can be added into the next iteration of combination. Notice that it is straightforward to prove that the combination of two triplet mass functions satisfies the associative property. The following theorem ensures the combinability of a collection of triplet mass functions.

**Theorem 4** Suppose we are given a collection of two-point mass functions $m_1, m_2, ..., m_n$ over the frame of discernment $\Theta$, then these mass functions are combinable if and only if any pair of $m_i$ and $m_j$ ($i \neq j$) is combinable.

Based on Theorem 4, we now consider general formulae for computing a collection of triplet mass functions. From Equation (2), a collection of triplet functions can be formulated on the basis of one focus being equal, two focuses being equal, and none of focuses being equal among triplets as follows:

$$m = \underbrace{m_{11} \oplus ... \oplus m_{1l_1}}_{l_1 \text{ terms}} ... \oplus \underbrace{m_{k1} \oplus ... \oplus m_{kl_k}}_{l_k \text{ terms}} \qquad (29)$$

where $1 \leq k \leq 3; 0 \leq l1, ..., l_k$ and $l_1 + ... + l_k = n$, and $n$ is the number of terms to be summed. For each $l_i$, we can apply the formulae established in the preceding sections to calculate the combinations of $n$ triplet mass functions.

To develop algorithms for computing Equation (29) and to examine its time complexity, we establish approximate formulae for combining a collection of triplet mass functions with one focus being equal, instead of the exact formulae because it is difficult or even impossible to establish exact ones. Similarly, we can provide approximate formulae for the other cases. All the computations for combining triplet mass functions can be achieved in linear time.

Given a collection of triplet mass functions be $m_1, ..., m_l$ defined on $\{x, y_1, \Theta\}, ..., \{x, y_l, \Theta\}$, where $y_1 \neq ... \neq y_l$, and $m_i(\{x\}) = p_i, m_i(\{y_i\}) = c_i, m_i(\Theta) = r_i; p_i + c_i + r_i = 1$, and $i = 1, 2, ..., l$. Now we first consider the combination of $m_1, m_2$, below: By using Equations (18), (19) and (20), we have

$$(m_1 \oplus m_2)(\{x\}) = K(m_1(\{x\})m_2(\{x\})$$
$$+ m_1(\{x\})m_2(\Theta) + m_1(\Theta)m_2(\{x\})),$$
$$(m_1 \oplus m_2)(\{y_1\}) = K m_1(\{y_1\})m_2(\Theta),$$
$$(m_1 \oplus m_2)(\{y_2\}) = K m_1(\Theta)m_2(\{y_2\}),$$
$$(m_1 \oplus m_2)(\Theta) = K m_1(\Theta)m_2(\Theta),$$

and for convenience, we use an alternative formula to calculate $K^{-1}$ below:

$$K^{-1} = \sum_{X \cap Y \neq \emptyset} m_1(X)m_2(Y) = m_1(\{x\})m_2(\{x\})$$
$$+ m_1(\{x\})m_2(\Theta) + m_1(\Theta)m_2(\{x\})$$
$$+ m_1(\{y_1\})m_2(\Theta) + m_1(\Theta)m_2(\{y_2\})$$

After applying the outstanding rule on the combined result, we assume that the first focal element is $\{x\}$ and the second element remains $\{y_1\}$, then we simplify $K^{-1}$ as

$$K^{-1} = \sum_{X \cap Y \neq \emptyset} m_1(X)m_2(Y) = m_1(\{x\})m_2(\{x\})$$
$$+ m_1(\{x\})m_2(\Theta) + m_1(\Theta)m_2(\{x\})$$
$$+ m_1(\{y_1\})m_2(\Theta)$$

Substituting $p_i, c_i, r_i$ for $m_i$, we then have the combined result of $m_1, m_2$ is as follows:

$$m_1 \oplus m_2(\{x\}) = K(p_1 p_2 + p_1(1 - P_2 - c_2) + p_2(1 - p_1 - c_1))$$
$$m_1 \oplus m_2(\{y_1\}) = K(c_1(1 - p_2 - c_2))$$
$$m_1 \oplus m_2(\Theta) = K((1 - p_1 - c_1)(1 - p_2 - c_2))$$

where

$$K^{-1} = (p_1 p_2 + p_1(1 - p_2 - c_2) + p_2(1 - p_1 - c_1))$$
$$+ (1 - p_1 - c_1)(1 - p_2 - c_2) + c_2(1 - p_2 - c_2)$$

Repeating the same process above, we can obtain the approximate formulae for combining $l$ triplet mass functions below:

$$m(\{x\}) = K(\prod_{i=1}^{l} p_i + \sum_{i=1}^{l}(1 - d_i) \prod_{j=1, j \neq i}^{l} p_j + \lambda) \quad (30)$$

$$m(\{y_i\}) = K(p_i \prod_{i=2}^{l}(1 - d_i)) \quad (31)$$

$$m(\Theta) = K(\prod_{i=1}^{l}(1 - d_i)) \quad (32)$$

where $\lambda$ is a constant and

$$k^{-1} = (\prod_{i=1}^{l} p_i + \sum_{i=1}^{l}(1 - d_i) \prod_{j=1, j \neq i}^{l}(p_j + \lambda) +$$
$$(p_i \prod_{i=2}^{l}(1 - d_i)) + (\prod_{i=1}^{l}(1 - d_i)) \quad (33)$$

These formulae are approximations to the exact formulae. At first glance, the computational effort of computing formula (31) requires $l + l \times l$ calculations, therefore its time complexity is about $O(l + l^2)$. For formulae (31)-(33), their time complexity is about $O(3l + l^2)$, i.e. $O(l^2)$. However, because each triplet only involves two singletons, the combination of two triplets also results in two singletons after applying the outstanding rule. Thus, the time complexity for computing all the cases is about $O(3 \times 2l) = O(2 \times 3l) = O(2n)$, i.e. $O(n)$. Our experimental results validate the estimate of this time complexity.

## 5  EVALUATION

To perform comparative analysis on combining dichotomous functions and combining triplet functions with respect to both efficiency and accuracy, we have implemented a set of algorithms for combining the two types of mass functions: dichotomous mass functions and triplet mass functions respectively, and carried out a range of experiments in the domain of text categorization.

For our experiments, we chose a benchmark data set − 20-newsgroup − a collection of text documents. It consists of twenty categories and the total number of documents is 20,000. We first divided the data set into a training data set and a testing set by using a $70\% - 30\%$ proportion, and then used our rough sets-based learning algorithm to generate ten classifiers (classification models) from the training data set, denoted by $R_0, R_1, ..., R_9$. Given a testing document from the testing data set, every classifier will produce an output in the form of a set of numeric scores − probabilities. Each of the scores indicates a probability of the document is assigned to a corresponding category. We model each list of the scores into a triplet and dichotomous structure as a piece of evidence. With the testing data set and ten classifiers, for each document, ten outputs will be generated in the forms of triplet and dichotomous, respectively. So the total number of the triplets or dichotomous generated is $10 \times 6,000 = 60,000$.

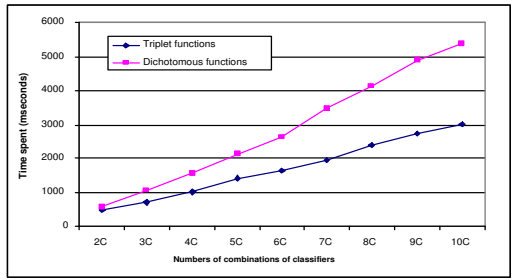

Figure 1: Efficiency comparison of combining dichotomous mass functions and combining triplet mass functions

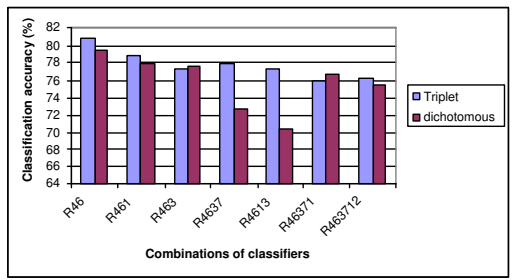

Figure 2: Comparison between classification accuracies of triplet and dichotomous

For the efficiency comparison, we first combine two lists of triplets from two classifiers, three lists of triplets, etc. until ten lists of triplets are combined. In the same way, we combine two lists of dichotomous, three lists of dichotomous, and so forth. Figure 1 presents a comparison of running time required in combining ten classifiers. The time required for combining the classifier outputs in the form of triplet is about 40% faster than that of dichotomous structure.

In order to compare classification accuracy between the two methods of triplet and dichotomous, we conduct further experiments. The first task is to experiment with various combinations of classifiers over 10 out of 20 document categories, such as combining two classifiers, three classifiers, etc. The second is to average the accuracies of the combined classifiers, and select some of the combined classifiers using the 70% accuracy as a cut-off point. Figure 2 depicts the performance of a group of the selected classifiers. The classification accuracy of combining triplet mass functions is 2.07% better than that of combining dichotomous mass functions on average.

## 6  SUMMARY

We have presented a theoretical basis validating our triplet structure and formulae for evidential reasoning. We have also established the general approximate formulae for combining evidential functions with the triplet structure. These formulae provide a theoretical proof that the time complexity of computing triplets can be achieved in linear-time and a basis for implementing the triplet-based algorithms. Apart from these, we have performed a comparative analysis to show the advantage of the triplet over the dichotomous with respect to the measure of basic probability assignment and efficiency in determining the final decisions.